\documentclass[11pt]{article}

\usepackage[preprint]{acl}

\usepackage{times}
\usepackage{latexsym}

\usepackage[T1]{fontenc}

\usepackage[utf8]{inputenc}

\usepackage{microtype}
\usepackage{amsmath, amssymb}   
\usepackage{array, multirow, boldline, makecell}
\usepackage{booktabs}
\usepackage[table]{xcolor}
\usepackage{inconsolata}

\usepackage{graphicx}

\usepackage[ruled,vlined]{algorithm2e}

\usepackage[most]{tcolorbox}
\usepackage{arydshln}
\usepackage{pifont}
\usepackage{soul}
\usepackage{float}
\usepackage{subfigure}
\usepackage{adjustbox}

\tcbset{colframe=black!80, colback=white, left=2pt, right=2pt, top=2pt, bottom=2pt, boxrule=0.6pt}

\title{Evontree: Ontology Rule-Guided Self-Evolution of Large Language Models}

\author{
  Mingchen Tu$^1$, 
  Zhiqiang Liu$^1$, 
  Juan Li$^1$, 
  Liangyurui Liu$^2$, 
  Junjie Wang$^3$, \\
  {\bf Lei Liang}$^3$, 
  {\bf Wen Zhang}$^{1}$\thanks{\hspace{1mm} Corresponding Author.} \\
  $^1$Zhejiang University, Hangzhou, China \\
  $^2$University of Electronic Science and Technology of China, Chengdu, China \\
  $^3$Ant Group, Hangzhou, China \\
  \texttt{\{mingchentz, zhiqiangliu, zhang.wen\}@zju.edu.cn}
}

\begin{document}
\maketitle
\begin{abstract}
Although Large Language Models (LLMs) perform exceptionally well in general domains, the problem of hallucinations poses significant risks in specialized fields such as healthcare and law, where high interpretability is essential. Existing fine-tuning methods depend heavily on large-scale professional datasets, which are often hard to obtain due to the privacy regulations. Moreover, existing self-evolution methods are primarily designed for general domains, which may struggle to adapt to knowledge-intensive domains due to the lack of knowledge constraints. In this paper, we propose an ontology rule guided method Evontree to enable self-evolution of LLMs in low-resource specialized domains. Specifically, Evontree first extracts domain ontology knowledge from raw models, then detects knowledge inconsistencies using two core ontology rules, and finally reinforces gap knowledge into model via self-distilled fine-tuning. Extensive evaluations on medical QA benchmarks using Llama3-8B-Instruct and Med42-V2 demonstrate the effectiveness of Evontree, which outperforms both the base models and strong baselines, achieving up to a 3.7\% improvement in accuracy. Detailed ablation studies further validate the robustness of our approach.

\end{abstract}

\section{Introduction}
Recently, large language models (LLMs) have demonstrated remarkable performance across a broad spectrum of domains, which is primarily attributed to extensive pre-training and subsequent fine-tuning on massive and specialized datasets. For instance, domain-specific models such as BioBERT \citep{biobert2020} and SciBERT \citep{scibert2019}, have shown that scaling pre-training to specialized corpora is crucial for superior downstream outcomes. More recent adaptation methods such as Med42-v2 \citep{christophe2024med42}, further underscore that instruction-tuning with billions of biomedical tokens remains a standard prerequisite for achieving optimal model performance.

However, acquiring such large-scale, high-quality domain data poses significant challenges in data-sensitive fields like healthcare and finance, due to stringent privacy regulations and high annotation costs. While existing domain fine-tuning methods depend heavily on external supervision, they often overlook the rich implicit ontology knowledge already internalized by LLMs during their pre-training phase \citep{DBLP:conf/emnlp/WangYXQD00GJX0C24}. Recent research has reached a consensus that LLMs can be viewed as implicit knowledge bases. Yet, without explicit constraints, these models frequently suffer from hallucinations and knowledge inconsistencies when applied to knowledge-intensive tasks. To bridge this gap, we introduce Ontology, a structured knowledge representation utilized by domain experts to describe concepts, relations, and attributes. 
Considering the ability to maintain logical integrity, experts define ontology rules to automatically check for consistency within a knowledge base. 
In low-resource scenarios, these structured rules offer a high-density form of supervision that is more accessible than massive raw corpora.

In this paper, we propose Evontree, an ontology rule guided framework designed for the self-evolution of LLMs in specialized domains with minimal human supervision. Rather than traditional fine-tuning that introduces vast amounts of external data, Evontree focuses on refining the model's internal knowledge. Our approach treats the LLM as a source of "raw" ontology, using expert rules to detect and correct its internal inconsistencies. Specifically, our framework consists of three main steps: (1) Knowledge Extraction: We explicitly extract the implicit subclass and synonym relationships embedded within the LLMs. (2) Consistency Detection: We introduce two core ontology rules (as shown in Table~\ref{tab:rules}) to identify logical inconsistencies within the extracted knowledge. (3) Knowledge Reinforcement: We re-inject the corrected and refined knowledge back into the model via self-distilled fine-tuning.

Experimental results demonstrate that Evontree consistently improves performance across major benchmarks. Notably, our method allows Llama3-8B-Instruct and the highly-tuned Med42-v2 to achieve significant accuracy gains (up to 3.7\%), outperforming baselines that rely on large-scale supervised corpora. This confirms that leveraging a limited set of high-quality ontology rules can be more effective for domain adaptation than simply increasing data volume.

Our main contributions are as follows:
\begin{itemize}
    \item We are the first to utilize ontology rules for low-resource LLM self-evolution. By leveraging LLM's internal knowledge and simple ontology rules, we adapt LLMs to diverse professional fields where high-quality data is scarce.
    \item We propose a novel framework that integrates the extraction of implicit knowledge, rule-based refinement, and self-distilled re-injection, enabling effective self-evolution without external supervision. 
    \item Experimental results demonstrate that Evontree outperforms post-training methods that rely on significantly larger supervised datasets, validating the effectiveness of our framework.
\end{itemize}

\begin{table}[t]
\centering
\resizebox{\linewidth}{!}{
\begin{tabular}{ccc}
\toprule
\textbf{ID} & \textbf{Premise} & \textbf{Conclusion} \\
\midrule
R1 & $(x,\text{SynonymOf},y) \wedge (y,\text{SubclassOf},z)$ & $\Rightarrow (x,\text{SubclassOf},z)$ \\
R2 & $(x,\text{SubclassOf},y) \wedge (y,\text{SubclassOf},z)$ & $\Rightarrow (x,\text{SubclassOf},z)$ \\
\bottomrule
\end{tabular}
}
\caption{Ontology Rules Used.}
\label{tab:rules}
\end{table}
\section{Preliminary}
\subsection{Ontology}
Ontology captures concepts, their inter-connections, and rules within a specific domain. Specifically, (1) \textbf{Concepts} represent entities or categories within a domain, such as "Cell" in a medical ontology. (2) \textbf{Relationships} define how concepts are inter-connected. The most important relationships in ontologies are Hyponymy (Is-subclass-of) and Synonymy (Is-synonym-of). 
Hyponymy represents a hierarchical and subclass relationship. Synonymy indicates that two concepts are semantically equivalent. For example, "Muscle Cell" is a subclass of "Cell", while "Muscle Cell" and "Muscle Fiber" are synonyms.
(3) \textbf{Axioms (Rules)}: Ontologies are equipped with built-in rules which enable automated reasoning and consistency checking within knowledge graphs. For example, if (Concept A, Is-subclass-of, Concept B) and (Concept B Is-subclass-of Concept C), then it logically follows that (Concept A, Is-subclass-of, Concept C). However, if the ontology also includes (Concept A, Is-Not-A-subclass-of, Concept C), a conflict arises with the inferred relationship. Such rules allow ontologies to automatically detect and resolve inconsistencies, ensuring the integrity of the knowledge graph. This capability is particularly valuable in large-scale knowledge bases, where manual verification is impractical.

\begin{figure*}[!htb]
\centering
\includegraphics[width=\textwidth]{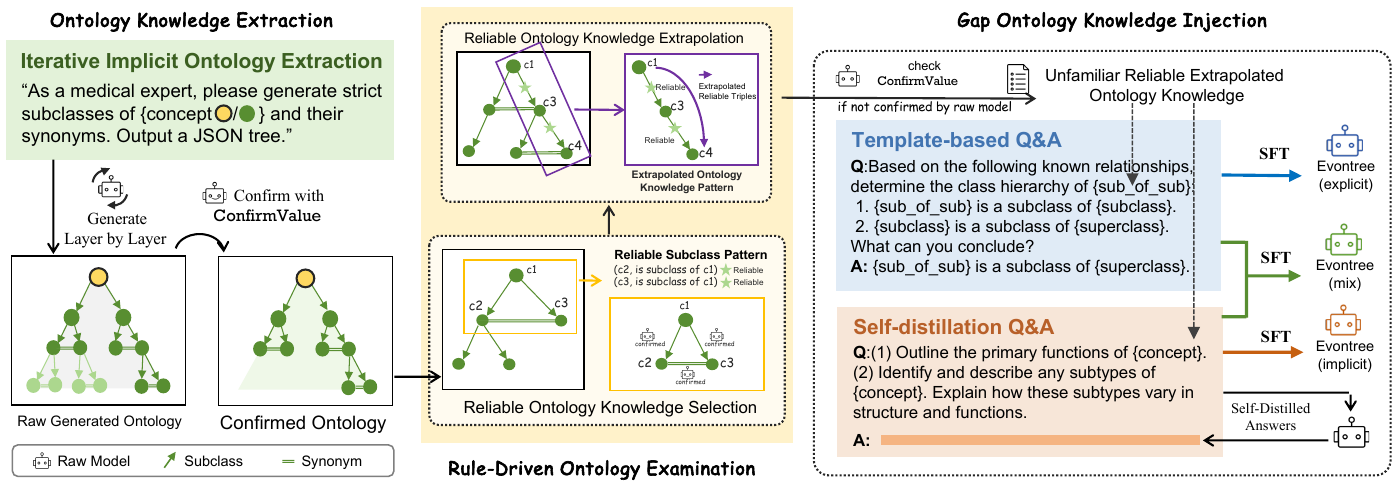}
\caption{The overview of Evontree.}
\label{fig:method}
\end{figure*}

\subsection{Perplexity}
 \label{sec:perplexity}
Model's perplexity plays a critical role in our calculation of metric $\mathtt{ConfirmValue}$. It quantifies the uncertainty of a probabilistic model in its predictions, where lower perplexity values correspond to higher prediction accuracy, while higher values indicate poorer performance. Formally, perplexity is defined as the exponential of the cross-entropy between the true distribution:
\begin{equation}
\text{Perplexity} = 2^{H(p, q)}
\end{equation}
where $H(p,q)$ is the cross-entropy between the true distribution $p$ and the model's predicted distribution $q$.
 In our framework, we leverage next-token prediction perplexity \citep{DBLP:journals/corr/abs-2403-09207,DBLP:conf/naacl/LiZLCC0W0024} to design our metric \textbf{$\mathtt{ConfirmValue}$}, which evaluates model's confirmation towards certain ontology triple relationship. 
 This helps mitigate hallucinations that arise from one-time generation.

\section{Methodology}
As illustrated in Figure ~\ref{fig:method}, our method comprises the following key steps: (1) Ontology Knowledge Extraction, (2) Rule-Driven Ontology Examination, and (3) Gap Ontology Knowledge Injection.

\subsection{Ontology Knowledge Extraction}\label{4.1}
\noindent \textbf{Raw Triples Extraction:} Firstly, we use 15 root concepts to initiate the ontology extraction process, including
Antibiotic, Bacterium, Cell, Enzyme, Fungus, Hormone, Tissue, Vertebrate, Virus, Vitamin, Chemical, Inorganic Chemical, Organic Chemical, Infectious Disease and Non-Infectious Disease.  In order to ensure a sufficient volume of gap ontology knowledge for fine-tuning , we restrict these root concepts' hierarchy depth$\geq3$.

During the raw generation, for each concept $c$ (initiated by root concept we manually designated, then iteratively replaced by generated last-layer concept), we feed the model the prompt as follows:
\begin{adjustbox}{max width=\textwidth, scale=0.9, center}
\begin{tcolorbox}[title=Generate Ontology Tree,
  colback=gray!10!white, colframe=gray!70!black,
  ]
As a medical expert, please generate strict subclasses of \{concept\} and their synonyms.  
Output a JSON tree like below:
\begin{lstlisting}[basicstyle=\ttfamily\small,breaklines=true,numbers=none,frame=single]
{"{c}": {
    "description": "",
    "subclasses": [{
        "name": "",
        "description": "",
        "synonyms": ["", ""]
}]}}
\end{lstlisting}
\end{tcolorbox}
\end{adjustbox}

After completing the iterative generation process, we obtain one ontology tree for each root concept, spanning at least three hierarchical layers. The initiated root concept is the apex of the ontology tree, and the underlying nodes are sub-concepts and synonym concepts generated layer-by-layer by the model itself. Edges of the tree denote the two most prevalent and fundamental ontology relations, subclass and synonym relationships.

\noindent\textbf{$\mathtt{ConfirmValue}$ Computation:} To reduce hallucination from one-time generation of models, we design the metrics $\mathtt{ConfirmValue}$, a causal-perplexity(introduced in Section~\ref{sec:perplexity})-based confidence score, to evaluate model's confirmation extent towards a single ontology triple. Taking raw triple $(A,\text{SynonymOf},B)$ as an example, we probe the model with the prompt as following.

\begin{adjustbox}{max width=\textwidth, scale=0.9, center}
\begin{tcolorbox}[title={Calculate ConfirmValue},
  colback=gray!10!white, colframe=gray!70!black]
Please determine if the statement is true or false, then answer with True or False: 'A' is an exact synonym of 'B'. Answer:
\end{tcolorbox}
\end{adjustbox}

For each triple $t$ under prompt $p$, we construct two completions:(1) $S_{\text{True}} = p + \text{``True''}$, (2) $S_{\text{False}} = p + \text{``False''}$, and compute the perplexity over the ``Answer: True/False'' token span. The $\mathtt{ConfirmValue}$ is defined as
\begin{equation}
    \mathtt{ConfirmValue}(t,p)=
\frac{\operatorname{sign}\bigl(\text{PPL}_{\text{False}}-\text{PPL}_{\text{True}}\bigr)}
     {\min\bigl(\text{PPL}_{\text{True}},\,\text{PPL}_{\text{False}}\bigr)}.
\end{equation}
To robustly gauge model’s confirmation towards certain ontology triple, we craft five paraphrased prompts for synonym relations and four for subclass relations. A triple is deemed model-confirmed only when every associated $\mathtt{ConfirmValue}$ surpasses a threshold $\tau$.

\noindent \textbf{Confirmation Threshold Setting:} The threshold $\tau$ is obtained for each model and each prompt. 
For each model–prompt pair, we regard the one-time generated output (true / false) as a binary decision signal and $\mathtt{ConfirmValue}$ as its continuous confidence score.
Using the $\approx$ 10 k raw subclass and synonym ontology triples from raw generation, we sweep $\tau \in [0,1]$ and compute the true-positive and false-positive rates. The optimal threshold $\tau^{\!*}$ is the cut-off that maximizes the Youden index $J=\mathrm{TPR}-\mathrm{FPR}$, i.e. the point where the binary decision and continuous confidence are best aligned. If the model designates certain ontology triple with $\mathtt{ConfirmValue}$ higher than threshold $\tau$, then we deem that this ontology triple is "confirmed" by the raw model. If the model assigns a $\mathtt{ConfirmValue}$ exceeding the threshold $\tau$ to a given ontology triple, we regard that triple as "confirmed" by the raw model.

\subsection{Rule-Driven Ontology Examination}\label{3.3}

\noindent\textbf{Reliable Ontology Knowledge Selection:}
While the preceding steps yield triples that are confirmed by the model, such confirmation alone does not ensure factual reliability. To guarantee that only robust knowledge is employed for downstream extrapolation, we must avoid arbitrarily selecting triples that may be erroneous. Therefore, we apply ontology rule R1 to identify closed triangles consisting of two subclass triples and one synonym triple, where the mutual structure allows each edge to corroborate the others. Any subclass triple that forms part of such a triangle is considered reliable. This rule leverages both ontological constraints and our prior belief that mislabelled instances within the model’s pre-training corpus are rare; therefore, the likelihood of three mutually supporting but erroneous statements co-occurring is extremely low. Consequently, only reliable subclass triples are retained for extrapolation in the subsequent step.

\noindent\textbf{Ontology Knowledge Extrapolation:}
We take the reliable subclass triples $\mathcal T_{\mathrm{rel}}$ from last step and apply ontology rule R2 to generate new extrapolated subclass triples $\mathcal T_{\mathrm{extrapolated}}$. For example, given reliable triples
\((D,\ \text{subClassOf},\ C)\) and \((C,\ \text{subClassOf},\ A)\),
rule~R1 yields \((D,\ \text{subClassOf},\ A)\).

\noindent \textbf{Gap Ontology Triples Selection:} 
For every $t_{\mathrm{extrapolated}}\!\in\!\mathcal T_{\mathrm{extrapolated}}$, we recompute its $\mathtt{ConfirmValue}$ using the same prompt set as the raw ontology extraction step.
Crucially, we retain only those triples whose $\mathtt{ConfirmValue}$ is below the threshold $\tau^{\!*}$ , which means those triples the model are not familiar with before.
These low-ConfirmValue triples are labelled $\mathcal T_{\text{gap}}$ , which are supposed to be injected into the model during the subsequent model's fine-tuning stage.

\subsection{Gap Ontology Knowledge Injection}

\noindent \textbf{Explicit Injection:} We introduce three injecting strategies for incorporating these credible but previously unfamiliar triples into the model. The simplest and most straightforward approach is to leverage our reliable and extrapolated triples to construct explicit reasoning chains, which are then used to synthesize question–answer pairs; the generation template is illustrated in the main figure~\ref{fig:method}. 

\noindent \textbf{Implicit Injection:} To mitigate homogeneity in synthesized QA data from single-prompt reasoning chains, we follow prior work \cite{OntoTune} to generate more natural and diverse training corpus. We append the ontology chain to pre-defined, concept-specific question templates in order to guide model to produce concept-aware higher-quality answers. We then finetune the model with the instruction and self-distilled output pairs back into the model, ensuring the ontology knowledge and concept-aware, self-distilled knowledge is reinforced to the model to continually enhance model's domain capability. 

The two pre-defined concept-centric question templates are shown in below.

\begin{adjustbox}{max width=\textwidth, scale=0.9, center}
\begin{tcolorbox}[title={Generate Corpus},
  colback=gray!10!white, colframe=gray!70!black]
(1) Outline the primary functions of \{concept\}.
(2) Identify and describe any subtypes of \{concept\}. Explain how these subtypes vary in structure and function.
\end{tcolorbox}
\end{adjustbox}

For each gap triple $(D,\ \text{SubclassOf},\ A)$ and its derivation chain $(D,\ \text{SubclassOf},\ C)$, $(C,\ \text{SubclassOf},\ A)$, we instantiate the placeholders with the three involved concepts $A,C,D$ for template (1), and with $A,C$ for template (2).

We hypothesise that appending the derivation chain as a \emph{hint} after each template enables the model to produce higher-quality answers.
The hint template are shown in following substantiated example. Given the weak triple (Skeletal Muscle Fiber,\ \text{SubclassOf},\ Cell) supported by two reliable chains:(i) (Skeletal Muscle Fiber,\ \text{SubclassOf},\ Muscle Cell), (Muscle Cell,\ \text{SubclassOf},\ Cell). (ii) (Skeletal Muscle Fiber,\ \text{SubclassOf},\ Myocyte), (Myocyte,\ \text{SubclassOf},\ Cell). 
we generate the following instruction for chain (ii):

\begin{adjustbox}{max width=\textwidth, scale=0.9, center}
\begin{tcolorbox}[title={Hint with Ontology Chain},
  colback=gray!10!white, colframe=gray!70!black]
{
  Outline the primary functions of Skeletal Muscle Fiber. You can consider these relationships as follows, but please ignore them if they are unnecessary: Skeletal Muscle Fiber is a subclass of Myocyte, and Myocyte is a subclass of Cell.
}
\end{tcolorbox}
\end{adjustbox}

The same process is repeated for every supporting chain.

\noindent \textbf{Mixed Injection}
As a further variant, we simultaneously inject knowledge via both implicit and explicit ways, which is termed mixed injection.

\section{Experiments}

We designed four research questions to verify the effectiveness of our method: (RQ1) Can our method accurately identify factually reliable ontology knowledge that the raw model fails to master? (RQ2) Can our method effectively improve raw model's domain capability without introducing any external corpora? (RQ3) How does Evontree impact raw model's generalization performance? (RQ4) Does every key component of Evontree make a meaningful contribution?

\subsection{Experiment Settings}
\label{ES}
\subsubsection{Datasets}
Following prior work~ \citep{OntoTune,gururajan2024aloe,christophe2024med42}, we adopt 3 widely-used medical datasets, including PubMedQA \citep{Jin_Dhingra_Liu_Cohen_Lu_2019}, MedQA \citep{medqa} and MedMCQA \citep{pal2022medmcqa} for evaluation. 

\subsubsection{Baselines}

We use Llama-3-8B-Instruct as our base model and additionally include Med42-v2\citep{christophe2024med42}, a Llama-3-8B variant fine-tuned in medical domains to test the effectiveness of our framework on an already strong domain model.

\subsubsection{Implementation}
Models are fine-tuned using LoRA via LLaMA-Factory\citep{zheng2024llamafactory} with a learning rate of $5e-5$ for 3 epochs. During the SFT stage, we use fp32 and a learning rate of 5e-5, training for 3 epochs with a cosine scheduler, a batch size per device initialized to 8 and gradient accumulation of 2. 

\subsubsection{Evaluation}
We evaluate the effectiveness of our approach from two perspectives. On the one hand, we employ GPT-4o-mini\citep{openai2023gpt4} and DeepSeek-V3\citep{deepseek2024v3} as strong supervisory models to assess the quality of the ontology triples generated at each stage of our pipeline. On the other hand, we measure our model’s performance on both medical-domain QA datasets and general capability benchmarks, following the evaluation protocols established in prior work~ \citep{OntoTune}.

\subsection{Characteristics of Ontology Triples (RQ1)}
\begin{figure}
    \centering
    \includegraphics[width=0.9\linewidth,height=4.5cm]{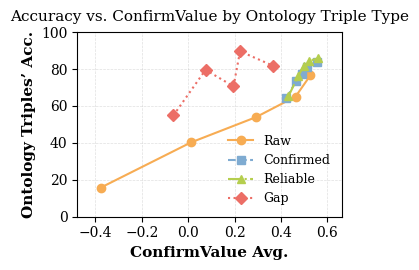}
    \caption{Relationship between accuracy and $\mathtt{ConfirmValue}$.}
    \label{fig:triple_acc}
\end{figure}

We observe the characteristics of ontology knowledge at different stages by analyzing the relationship between their factual accuracy and the model's self-confidence ($\mathtt{ConfirmValue}$), as illustrated in Figure~\ref{fig:triple_acc}.

\textbf{Effectiveness of Reliable Ontology Knowledge Selection.} A direct comparison between \textit{Confirmed triples} (blue) and \textit{Reliable triples} (green) validates the effectiveness of our \textbf{Reliable Ontology Knowledge Selection} module. While both types occupy a similar range of high $\mathtt{ConfirmValue}$, the reliable triples consistently achieve higher accuracy across the spectrum. This demonstrates that our unsupervised selection strategy successfully identifies more factually accurate knowledge beyond what simple confidence thresholds can provide.

\textbf{Identifying Knowledge Gaps.} The contrast between \textit{Gap triples} (red) and the other categories (blue and green) reveals a critical finding. Although the accuracy of gap triples is comparable to confirmed and reliable ones, their average $\mathtt{ConfirmValue}$ is significantly lower. This pattern aligns perfectly with our core hypothesis: Evontree effectively isolates "blind spots"—knowledge that is factually correct but for which the raw model lacks internal certainty. By targeting these low-confidence but high-accuracy triples, we ensure that the self-evolution process focuses on the most valuable and reliable information gaps.

\begin{table}[]
      \resizebox{\linewidth}{!}{
        \begin{tabular}{lcccc}
          \toprule
          \textbf{Triple Type} & \textbf{Relation} & \textbf{Num} & \textbf{$\mathtt{ConfirmValue}$ Avg.} & \textbf{Acc.} \\
          \midrule
          Raw  & synonym  & 95{,}814  & 0.0537 & 54.90 \\
          Confirmed  & synonym  & 30{,}532  & 0.4738 & 87.51 \\
          \midrule
          Raw   & subclass  & 58{,}309  & 0.2094 & 47.79 \\
          Confirmed  & subclass & 24{,}276  & 0.4896 & 72.62 \\
          Reliable  & subclass & 14{,}169  & 0.4981 & 74.81 \\
          Extrapolated  & subclass & 8{,}661  & 0.4449 & 80.29 \\
          Gap & subclass & 1{,}396   & 0.1595 & 74.57 \\
          \bottomrule
        \end{tabular}
      }
      \caption{Ontology path statistics and quality evaluation.}
      \label{tab:ontology-path-eval}
    \end{table}

\subsection{Evaluation on Medical Datasets (RQ2)}
As shown in Table \ref{tab:main}, among the three Evontree variants, the ``explicit'' ontology injection approach does not effectively improve model performance, whereas the more natural ``implicit'' and ``mix'' variants yield significant enhancements. This trend is further validated when comparing TaxoLLaMA (which uses explicit ontology injection) and OntoTune (which adopts implicit injection). Notably, Evontree not only achieves substantial performance gains for general-purpose LLMs (e.g., LLaMA3-8B-Instruct), but also further boosts domain-specific models like Med42-v2, reaching improvements of 3.1\% and 3.7\%, respectively. Compared with existing LLaMA3-8B-based domain models, Evontree attains state-of-the-art results even without introducing additional external supervision data, outperforming baselines by an average of 1.1\%. This indicates that large-scale, fragmented corpora alone cannot reliably achieve effective domain alignment, and may even introduce conflicting knowledge that increases model confusion. We conclude that the advantage of Evontree lies in its targeted identification of effective gap triples within the model's knowledge system. Through fine-grained automated selection, Evontree ensures that only information highly compatible with the model's existing knowledge structure is injected. This mechanism enables the model to self-reflect and precisely locate knowledge gaps, evolving itself based on reliable information while effectively avoiding unnecessary noise and conflicts. Compared to traditional data augmentation and domain adaptation approaches, Evontree prioritizes quality over quantity in knowledge supplementation, promoting greater consistency and completeness in the model's knowledge base.

\begin{table*}[t!]
\resizebox{\textwidth}{!}{
\begin{tabular}{clcccc>{\columncolor{gray!15}}c}
\toprule
\textbf{Setting} & \textbf{Model} & \textbf{MedQA} & \textbf{MedMCQA} & \textbf{PubMedQA} & \cellcolor{gray!15} \textbf{Average} \\
\midrule
& LLaMA3 8B~ \citep{llama3-8b_2024} & 51.7 & 51.7 & 70.3 & \cellcolor{gray!15}57.9 \\
&TaxoLLaMA~ \citep{DBLP:journals/corr/abs-2403-09207} & 50.5 & 46.1 & 73.4 & \cellcolor{gray!15}56.7 \\
& OntoTune$_{sft}$~ \citep{OntoTune} & 51.5 & 56.7 & 72.0 & \cellcolor{gray!15}\underline{60.1} \\
& OntoTune$_{dpo}$~ \citep{OntoTune} & \textbf{53.3} & \underline{57.2} & 65.5 & \cellcolor{gray!15}58.7  \\
{zero-shot}
& OntoTune$_{sft+dpo}$~ \citep{OntoTune} & 51.9& 56.7 & 66.3 & \cellcolor{gray!15}58.3  \\  
  \cmidrule{2-6}
& LLaMA3 8B-Evontree (explicit)& 43.4& 48.6 & \textbf{76.4} & \cellcolor{gray!15}56.1  \\ 
& LLaMA3 8B-Evontree (implicit) & \underline{52.7} & 54.4 & 72.9 & \cellcolor{gray!15}60.0  \\ 
& LLaMA3 8B-Evontree (mix) & 51.0& \textbf{57.4} & \underline{74.7} & \cellcolor{gray!15}\textbf{61.0}  \\
\cmidrule{2-6}
& $\Delta$ {Improvement over raw model} & +1.0\% & +5.7\% & +6.1\% & \cellcolor{gray!15}+3.1\%  \\ 
& $\Delta$ {Improvement over best baseline} & -0.6\% & +0.2\% & +3.0\% & \cellcolor{gray!15}+0.9\% \\ 
\midrule
& LLaMA3 8B~ \citep{llama3-8b_2024} & 56.4 & 53.9 & 77.2 & \cellcolor{gray!15}62.5 \\
 & Aloe~ \citep{gururajan2024aloe} & 51.1 & 56.8 & 75.4 & \cellcolor{gray!15}61.1 \\
 & Med42-v2~ \citep{christophe2024med42} & 57.8 & 58.1 & 74.6 & \cellcolor{gray!15}63.5 \\
&TaxoLLaMA~ \citep{DBLP:journals/corr/abs-2403-09207} & 55.9 & 57.5 & 77.6 & \cellcolor{gray!15}63.7 \\
 & OntoTune$_{sft}$~ \citep{OntoTune} & \underline{58.4} & 60.4 & 78.6 & \cellcolor{gray!15}65.8 \\
 {SFT} & OntoTune$_{dpo}$~ \citep{OntoTune} & 58.3 & \underline{60.7} & \textbf{79.4} & \cellcolor{gray!15}\underline{66.1} \\
(on evaluation)
&OntoTune$_{sft+dpo}$~ \citep{OntoTune} & 58.2 & 60.5 &\underline{78.9} & \cellcolor{gray!15}65.9 \\
\cmidrule{2-6}
& Med42-v2-Evontree (explicit)& 57.1& 58.6 & 71.1 & \cellcolor{gray!15}62.3  \\ 
& Med42-v2-Evontree (implicit) & \textbf{60.3} & \textbf{62.4} & \underline{78.9} & \cellcolor{gray!15}\textbf{67.2}  \\ 
& Med42-v2-Evontree (mix) & 57.2& 57.7 & 74.9 & \cellcolor{gray!15}63.3  \\ 
\cmidrule{2-6}
& $\Delta$ {Improvement over raw model} & +2.5\% & +4.3\% & +4.3\% & \cellcolor{gray!15}+3.7\%  \\ 
& $\Delta$ {Improvement over best baseline} & +1.9\% & +1.7\% & -0.5\% & \cellcolor{gray!15}+1.1\% \\ 
\bottomrule
\end{tabular}}
\centering
\caption{Results of the medical domain QA in the zero-shot and supervised fine-tuning (on evaluation) setting. The best results are highlighted in bold, while the second best are underlined. Results are taken from OntoTune~ \citep{OntoTune}, and we evaluate our method in the same way.}
\label{tab:main}
\end{table*}

\begin{table*}[ht]
\resizebox{\textwidth}{!}{
\begin{tabular}{lccccc|cc|c|cc}
\toprule
\multirow{2}{*}{\textbf{Model}} & \multicolumn{5}{c}{\textbf{MMLU}} & \multicolumn{2}{c}{\textbf{ARC}}  & \multicolumn{1}{c}{\textbf{TriviaQA}} & \multicolumn{2}{c}{\textbf{Advbench}}       \\ 
\cmidrule{2-11}
& \textbf{STEM} & \textbf{Social Sciences} & \textbf{Humanities} & \textbf{Other} & \textbf{Average} & \textbf{ARC\_C} & \textbf{ARC\_E} & \textbf{-} & \textbf{Raw Safe} & \textbf{Jailbreak Safe} \\ 
\midrule
LLaMA3 8B~ \citep{llama3-8b_2024} & 56.83 & 76.61 & 60.81 & 74.10 & 66.49 & 78.64 & 92.77 & 64.81 & 97.50 & 96.35 \\
TaxoLLaMA~ \citep{DBLP:journals/corr/abs-2403-09207} & 55.96 &73.74 &56.92 &69.43 &63.29 &72.88  &89.24  &63.12 &94.04 &73.27 \\
OntoTune$_{sft}$~ \citep{OntoTune} &56.47 &75.73 &\textbf{61.85} &\underline{73.02} &\textbf{66.31} & 78.31& \underline{91.89}  &\textbf{64.07}&94.04&\underline{92.69} \\
OntoTune$_{dpo}$~ \citep{OntoTune} &56.33 &75.33 &59.93 &\textbf{73.64} &65.70 & 78.98 &\textbf{92.06} &\underline{63.96} &90.58 &77.88 \\
OntoTune$_{sft+dpo}$~ \citep{OntoTune} &55.67 &75.17 &\underline{61.79} &72.71 &65.93 &78.98 &\textbf{92.06} &\underline{63.96} &90.58 &84.81 \\
LLaMA3-Evontree (implicit) &\underline{56.89} & 76.41 & 61.38 & 71.99 & 66.16 & 79.66 & 89.59 & 63.00 & \underline{95.38} & \textbf{96.15} \\
LLaMA3-Evontree (mix) & \textbf{56.99} & \textbf{76.80} & 61.62 & 71.71 & \underline{66.28} & \underline{80.34} & 90.30 & 63.22 &95.19 &\textbf{96.15} \\
LLaMA3-Evontree (explicit) &\underline{56.89} & \underline{76.60} & 60.85 & 72.24 &66.08 & \textbf{82.03} & 91.71 & 63.42 & \textbf{97.50} & 83.65 \\

\midrule
Aloe ~ \citep{gururajan2024aloe} & 55.67 & \underline{76.24} & 58.91 & 72.25 & 65.10 & 75.25& 86.95  &63.03 & 62.50 &34.23 \\
Med42-v2~ \citep{christophe2024med42} & 56.59 & \underline{76.24}& 59.91& 72.67& 65.72& \textbf{82.37} & \textbf{92.59} &\textbf{65.19}& 83.85& 60.19 \\
Med42-v2-Evontree (implicit)	&\textbf{56.86}	&\textbf{76.37}	&59.79	&72.79	&\underline{65.80}	&80.34	&91.36	&63.54	&\underline{86.54}	&\underline{60.77} \\
Med42-v2-Evontree (mix) 	&56.46	&\underline{76.24}	&\textbf{60.09}	&\underline{72.86}	&\underline{65.80}	&\underline{80.68}	&\underline{92.06}	&63.71	&\textbf{89.23}	&\textbf{65.77} \\
Med42-v2-Evontree (explicit)	&\underline{56.79}	&76.11	&\underline{59.98}	&\textbf{72.92}	&\textbf{65.82}	&\underline{80.68}	&91.18	&\underline{64.33}	&75.38	&45.19 \\
\bottomrule
\end{tabular}}
\centering
\caption{Results of general capabilities and safety evaluation.The results with the least decrease compared to the original model are highlighted in bold, while the second least are underlined. }
\label{tab:general}
\end{table*}

\subsection{General Capabilities and Safety Evaluation (RQ3)}
\noindent\textbf{General Capabilities Evaluation.} We evaluate the general capabilities of our model variants compared to the raw model on the MMLU\citep{hendryckstest2021mmlu}, TriviaQA\citep{joshi2017triviaqa}, and ARC datasets. Specifically, MMLU is assessed using the LLaMAFactory framework, while ARC and TriviaQA are evaluated using the OpenCompass tool in gen mode. As shown in Table ~\ref{tab:general}, under the zero-shot setting, all three variants of our method exhibit no significant degradation in general capabilities. The best-performing variant shows an average performance drop of only 0.15\% compared to the raw model, and LLaMA3-Evontree(mix) even achieves marginal improvements on certain test sets.Furthermore, under the supervised fine-tuning (SFT) evaluation setting, our models continue to demonstrate similar trends. Compared to med42-v2, our variants maintain stable performance and in some cases achieve notable gains, highlighting the robustness and potential benefits of our approach.

\noindent\textbf{Safety Evaluation.} Following prior work, we also assess whether our method introduces any safety risks to the model. We use harmful instructions from the AdvBench dataset to measure the proportion of safe responses, reported as the "Raw Safe" metric. Then, we append an inducing suffix to the harmful prompts to encourage unsafe behavior and measure the proportion of safe outputs under this adversarial setting, reported as the "Jailbreak Safe" metric. As shown in Table ~\ref{tab:general}, the Evontree(implicit) and Evontree(mix) variants show no significant drop in safety compared to the raw model, and in fact, when built upon Med42, these two variants even lead to notable improvements. In contrast, the Evontree(explicit) variant performs worse in both the LLaMA3 and Med42-v2 settings. We hypothesize that, similar to the observations in TaxoLLaMA, the introduction of rigid template-based Q\&A formats may negatively impact the model’s safety performance.

\subsection{Ablation Study (RQ4)}
\begin{table}[]
\centering
      \resizebox{0.8\linewidth}{!}{
      \centering
        \begin{tabular}{lccc}
          \toprule
          \textbf{Variant} & \textbf{gpt} & \textbf{deepseek} & \textbf{Avg.} \\
          \midrule
          w/o reliable triple sel. & 58.4 & 53.0 & 55.7 \\
          Evontree & 75.3 & 73.8 & 74.6 \\
          \bottomrule
        \end{tabular}
      }
      \caption{Impact of Reliable-Triple Selection on Extrapolated Triple Accuracy.}
      \label{tab:w/o reliable triple selection triple eval}
    \end{table}

    \begin{table}[]
      \resizebox{\linewidth}{!}{
        \begin{tabular}{ccccc}
          \toprule
          \textbf{Variant} & \textbf{MedQA} & \textbf{MedMCQA} & \textbf{PubMedQA} & \cellcolor{gray!20}\textbf{Avg.} \\
          \midrule
          w/o reliable triple sel.     
              & \underline{52.4} 
              & \textbf{54.7} 
              & 72.3 
              & \cellcolor{gray!20}\underline{59.8} \\
          w/o gap triple sel.          
              & 52.6 
              & 54.1 
              & \underline{72.5} 
              & \cellcolor{gray!20}59.7 \\
          w/o ontology injecting            
              & 51.1 
              & 53.9 
              & \textbf{73.1} 
              & \cellcolor{gray!20}59.4 \\
          LLaMA3-Evontree(implicit)         
              & \textbf{52.7} 
              & \underline{54.4}
              & \underline{72.9} 
              & \cellcolor{gray!20}\textbf{60.0} \\
          \bottomrule
        \end{tabular}
      }
      \caption{Ablation Study of Different Modules.}
      \label{tab:ablation}
    \end{table}

\begin{figure}
\centering
\includegraphics[scale=0.67]{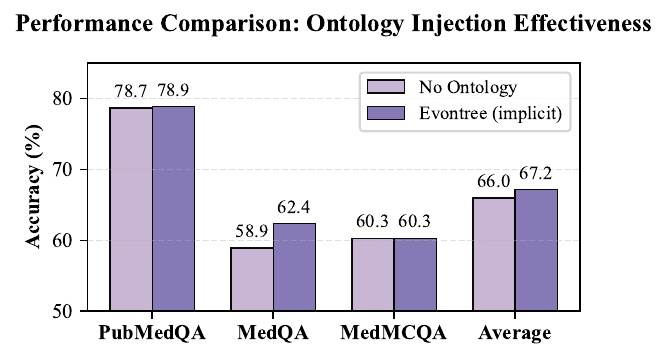}
\caption{Ablation Study of Med42-v2.}
\label{fig:ontology_injecting}
\end{figure}

The ablation results in Table \ref{tab:ablation} and Figure \ref{fig:ontology_injecting} demonstrate that Evontree’s performance stems from the synergy of its three core modules:
(1)\textbf{Ablation on Reliable Ontology Knowledge Selection.} The significant accuracy gap in Table \ref{tab:w/o reliable triple selection triple eval} (74.6\% vs. 55.7\%) underscores that simple confidence scores ($\mathtt{ConfirmValue}$) are insufficient for identifying truth, justifying our rule-based reliable selection. 
(2) \textbf{Ablation on Gap Triple Selection.} The performance drop across all three medical benchmarks when removing Gap Triple Selection (Table \ref{tab:ablation}) confirms that the model benefits most from knowledge it is "uncertain" about, rather than redundant information.
(3)\textbf{Ablation on Ontology Injecting.} The fact that the "w/o ontology injecting" variant declined the most indicates that structured ontology serves as a vital anchor for reasoning, preventing the model from reverting to hallucinated patterns during self-distillation. 

Collectively, these ablation studies confirm that Evontree's effectiveness stems from its ability to transform implicit, unverified model knowledge into explicit, factually-grounded domain expertise.

\section{Related Work}
\subsection{Domain Adaptation and Knowledge Integration}
Large Language Models (LLMs) often require specialized adaptation to perform effectively in knowledge-intensive fields \citep{cont-pretra2024}. Early efforts, such as BioBERT \citep{biobert2020} and BloombergGPT \citep{BloombergGPT2023}, relied on extensive pre-training on domain-specific corpora. More recent approaches like Med42-v2 \citep{christophe2024med42} and Aloe \citep{gururajan2024aloe} focus on instruction-tuning using billions of tokens. However, as noted in recent studies, acquiring such large-scale professional datasets is often hindered by privacy regulations and high annotation costs in sectors like healthcare and finance.

To mitigate hallucinations and improve reasoning, ontology-integrated methods have emerged to align LLMs with structured knowledge. Frameworks such as TaxoLLaMA \citep{DBLP:journals/corr/abs-2403-09207} and OntoTune \citep{OntoTune} incorporate external, human-curated ontologies (e.g., SNOMED CT) to provide semantic constraints.

In contrast to these data-dependent paradigms, our approach recognizes that LLMs already function as implicit knowledge bases \citep{DBLP:conf/emnlp/WangYXQD00GJX0C24}. Rather than injecting external supervision, we focus on extracting and refining the model's internal ontology knowledge, achieving robust domain adaptation in low-resource scenarios without relying on any external annotated corpora.

\subsection{Self-Evolution of LLMs}
Self-evolution allows LLMs to autonomously generate training data, reducing the need for human labor. For instance, SELF-INSTRUCT \citep{self-instruct2023} enables models to bootstrap task instructions and responses. To ensure the quality of self-generated content, various selection heuristics prioritize logical consistency \citep{self-improve2023, self-consistency2023} or adhere to predefined ethical criteria \citep{self-alignment2023}. Additionally, Chain-of-Thought (CoT) distillation \citep{distill2024} is frequently used to enhance the reasoning depth of generated outputs.

However, existing self-evolution methods are primarily designed for general domains and often struggle in specialized fields due to a lack of rigorous knowledge constraints. Unlike these general-purpose heuristics, our framework, \textbf{Evontree}, introduces knowledge-related ontology rules as a  supervision, enabling effective self-evolution in specialized domains.

\section{Conclusion}
In this work, we address the challenge of adapting large language models to data-scarce domains by leveraging the unique value of domain ontology rules. Our framework enables explicit extraction and validation of LLMs' implicit knowledge, using only two carefully chosen ontology rules to systematically detect and correct inconsistencies. By re-injecting revised knowledge into the model, we substantially improve performance on medical QA tasks, outperforming baselines dependent on large supervised datasets. Extensive experiments across two representative models and several medical benchmarks validate the robustness and effectiveness of our paradigm. This study highlights the potential of ontology rule-driven supervision as a practical and powerful solution for enhancing LLMs in highly specialized domains where conventional data-centric approaches are limited by privacy or scarcity. Future work can explore broader application of our framework to other professional domains and further enrichment of ontology rule-guided knowledge editing techniques.

\section{Limitations}
Despite its effectiveness, Evontree has several limitations that offer avenues for future work.

First, the framework currently requires 15 manually selected root concepts to initiate the ontology generation tree. While this is the only external input beyond our two rules, future research could develop a more automated "from-scratch" approach by leveraging self-evaluation and model consistency to identify adequate root concepts without human intervention.

Second, our experiments primarily focused on subclass and synonym relationships. While these are fundamental to most domain ontologies, the logic of Evontree is naturally extensible. Future work will explore more complex relationship types and diverse ontology rules, aiming to adapt our self-evolution framework to a broader range of specialized knowledge-intensive domains.


\bibliography{custom}

\end{document}